\documentclass[10pt,twocolumn,letterpaper]{article}

\usepackage{cvpr}              
\definecolor{cvprblue}{rgb}{0.21,0.49,0.74}
\usepackage[pagebackref,breaklinks,colorlinks,allcolors=cvprblue]{hyperref}
\usepackage{multirow}
\usepackage{array}
\usepackage{amsmath}
\usepackage[table]{xcolor}
\usepackage{multirow}
\usepackage{array}

\def\paperID{*****} 
\def\confName{CVPR}
\def\confYear{2026}

\title{\includegraphics[height=1em]{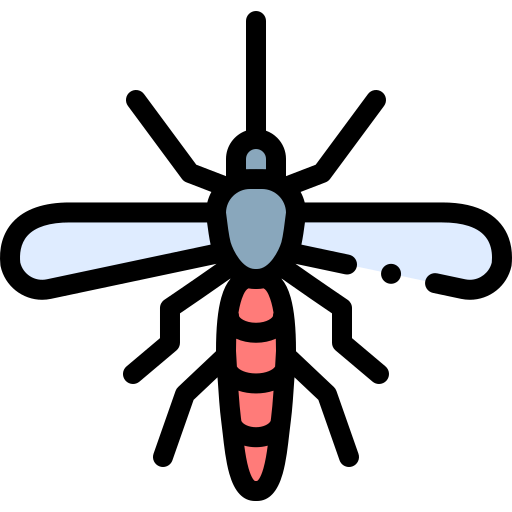} \textsc{VisText-Mosquito}: A Unified Multimodal Dataset for Visual Detection, Segmentation, and Textual Explanation on Mosquito Breeding Sites}

\author{
Md. Adnanul Islam$^{1, \dagger}$ \quad 
Md. Faiyaz Abdullah Sayeedi$^{1,3, \dagger}$ \quad 
Md. Asaduzzaman Shuvo$^{1, \dagger}$ \\
Shahanur Rahman Bappy$^{1, \dagger}$ \quad 
Muhammad Ziaur Rahman$^{1, \dagger}$ \quad 
Md Asiful Islam$^{2}$ \quad 
Swakkhar Shatabda$^{3}$ \\
\vspace{0.2cm}
\small $^{1}$United International University, Bangladesh \quad $^{2}$University of Arizona, USA \quad $^{3}$BRAC University, Bangladesh \\
\small \texttt{\{mislam221096, msayeedi212049, ashuvo221104, sbappy211002, mrahman202004\}@bscse.uiu.ac.bd} \\
\small \texttt{asifulislam@arizona.edu, swakkhar.shatabda@bracu.ac.bd} \\
\small $^{\dagger}$Equal contribution
}

\begin{document}
\maketitle
\begin{abstract}
Mosquito-borne diseases pose a major global health risk, requiring early detection and proactive control of breeding sites to prevent outbreaks. In this paper, we present \textsc{VisText-Mosquito}, a multimodal dataset that integrates visual and textual data to support automated detection, segmentation, and explanation for mosquito breeding site analysis. The dataset includes 1,828 annotated images for object detection, 142 images for water surface segmentation, and natural language explanation texts linked to each image. The YOLOv9s model achieves the highest precision of 0.92926 and mAP@50 of 0.92891 for object detection, while YOLOv11n-Seg reaches a segmentation precision of 0.91587 and mAP@50 of 0.79795. For textual explanation generation, we tested a range of large vision-language models (LVLMs) in both zero-shot and few-shot settings. Our fine-tuned \textbf{Mosquito-LLaMA3-8B} model achieved the best results, with a final loss of 0.0028, a BLEU score of 54.7, BERTScore of 0.91, and ROUGE-L of 0.85. This dataset and model framework emphasize the theme \textit{``Prevention is Better than Cure''}, showcasing how AI-based detection can proactively address mosquito-borne disease risks. The dataset and implementation code are publicly available at GitHub: \url{https://github.com/adnanul-islam-jisun/VisText-Mosquito}

\end{abstract}
    
\section{Introduction}
\label{sec:intro}

Mosquito-borne diseases continue to be a leading cause of illness and death worldwide, particularly affecting low- and middle-income countries. According to the World Health Organization (WHO), approximately 700 million people are affected by mosquito-borne illnesses every year, resulting in over one million deaths \cite{R2}. Diseases such as Malaria, Dengue Fever, Zika Virus, and Chikungunya are transmitted by mosquito vectors that breed primarily in stagnant water \cite{saleeza2013mosquito}. The economic burden associated with these diseases is also substantial; for instance, malaria alone is estimated to cost African economies over \$12 billion annually due to healthcare expenses and lost productivity \cite{R3}.

Despite ongoing efforts to combat these diseases, the incidence of dengue has increased 30-fold in the past 50 years, with nearly 390 million cases reported annually \cite{R4}. The rapid pace of urbanization and poor water management in many regions further exacerbate the problem by creating ideal conditions for mosquito breeding \cite{baitharu2021environmental}. Traditional mosquito control methods, including manual inspection and elimination of breeding sites, are labor-intensive, time-consuming, and often infeasible in large or inaccessible areas \cite{han2016novel}. These challenges underscore the urgent need for scalable, technology-driven approaches to detect and manage mosquito habitats more efficiently.

Although computer vision have made significant strides, the task of accurately detecting mosquito breeding places and analyzing water surfaces remains challenging \cite{bravo2021automatic}. Existing models often underperform due to a lack of specialized datasets capable of handling both detection and segmentation tasks \cite{R5}. Moreover, most datasets are unimodal, lacking contextual explanations that could aid in human interpretation or decision-making \cite{10328369}. This absence of interpretability limits the utility of AI models in real-world public health scenarios, where both performance and explainability are critical.


To address these gaps, we present \textsc{VisText-Mosquito}, the first multimodal benchmark for mosquito breeding site analysis. Our dataset integrates three complementary components: (i) object detection with 1,828 images and 3,752 bounding boxes across five container types (\texttt{coconut\_exocarp}, \texttt{vase}, \texttt{tire}, \texttt{drain\_inlet}, and \texttt{bottle}); (ii) water surface segmentation with 142 high-resolution images and 253 pixel-level masks; and (iii) natural language explanation with automatically generated and human-validated explanations that capture visual cues linked to mosquito breeding risk. Beyond dataset creation, we perform a comprehensive benchmarking study of state-of-the-art detectors, segmenters, and large vision–language models. Finally, we fine-tune \texttt{LLaMA3-8B-Vision} on our explanation corpus, yielding \texttt{Mosquito-LLaMA3-8B}, which achieves state-of-the-art BLEU, ROUGE-L, and BERTScore, demonstrating the value of domain adaptation for multimodal understanding. Our key contributions are as follows:

{\flushleft (1)} We introduce \textsc{VisText-Mosquito}, the first multimodal dataset that unifies object detection, water segmentation, and visual–linguistic explanation for mosquito breeding site analysis.
{\flushleft (2)} We provide high-quality annotations across visual and textual modalities, including bounding boxes, segmentation masks, and textual explanations for explainable AI, which are all human-annotated and validated.
{\flushleft (3)} We benchmark a wide spectrum of detectors, segmenters, and VLMs under zero-shot and few-shot setups, revealing the limitations of unimodal approaches.
{\flushleft (4)} We develop \texttt{Mosquito-LLaMA3-8B}, a domain-adapted VLM that sets a new baseline for multimodal reasoning in public health surveillance.
\section{Related Works}
\label{sec:related_works}

Mosquito-borne diseases continue to pose serious public health threats globally, especially in tropical and subtropical regions \cite{who2022mosquito}. Effective surveillance of mosquito breeding habitats is essential for early intervention and disease prevention. In recent years, advances in computer vision and machine learning have enabled automated detection of mosquito habitats using images and spatial data.

Several datasets have been proposed to support AI-based mosquito breeding site detection. The MosquitoFusion dataset \cite{R5} offers annotated real-world images capturing mosquitoes, larvae, swarms, and breeding sites across diverse contexts. Mehra et al. \cite{R6} created an entomological image dataset for training lightweight object detection models under challenging lighting and occlusion conditions. Similarly, Chathura et al. \cite{R7} presented a drone-captured image dataset that includes standing water sources and vegetation typically associated with mosquito breeding, enhancing the generalizability of deep learning models to outdoor environments. Object detection architectures such as YOLOv5, YOLOv8, and EfficientDet have shown promising results in localizing potential breeding containers, with YOLOv8 achieving a mean average precision (mAP@50) of 57.1\% on the MosquitoFusion dataset \cite{R5}. For semantic and instance segmentation, CNN-based models like U-Net and DeepLabV3+ have been used to segment water surfaces from drone and satellite imagery with pixel-level accuracy \cite{R6, A4}. Recent transformer-based models like DETR \cite{R8} and the Segment Anything Model (SAM) \cite{kirillov2023segment} offer superior segmentation quality by learning rich visual features from complex environments.

Geospatial AI techniques have been applied to UAV and satellite imagery for spatial mapping of mosquito breeding hotspots. For instance, GIS-based models using remote sensing data and LIDAR have enabled high-resolution risk mapping of urban habitats \cite{bravo2021automatic, A2}. Studies have also explored topographical features, vegetation indices, and hydrological data to predict stagnant water accumulation areas likely to host larvae \cite{R9}. Moreover, ensemble machine learning methods (e.g., Random Forest, XGBoost) using environmental and epidemiological data have improved mosquito vector prediction models, especially when coupled with weather and public health surveillance data \cite{R10}. In the multimodal domain, progress has been slower. Most existing works focus solely on image-based detection or segmentation, with limited incorporation of textual or reasoning-based outputs. However, natural language generation (NLG) in vision-language tasks has gained popularity, especially with models like BLIP \cite{li2022blip}, LLaVA \cite{liu2023llava}, and Flamingo \cite{alayrac2022flamingo}, which generate human-like captions or justifications. These models have been applied to medical imaging, environmental monitoring, and accessibility applications, showing promise for tasks requiring explainability and human alignment.

Despite these advancements, notable limitations persist. Many datasets are restricted to lab environments or synthetic scenes, lacking the visual complexity of urban and peri-urban breeding sites. Furthermore, most existing models treat breeding site identification as a visual-only task, ignoring the value of language-based explanations for public health decision-making and education. To the best of our knowledge, no prior work has introduced a multimodal dataset that jointly supports object detection, water surface segmentation, and visual explanation tailored to mosquito breeding site analysis. This represents a significant research gap, as integrating vision and language could boost both interpretability and field applicability of AI models for mosquito vector control.
\section{Dataset Construction Process}
\label{sec:dataset}

\begin{figure*}[t!]
    \centering
    \includegraphics[width=\textwidth]{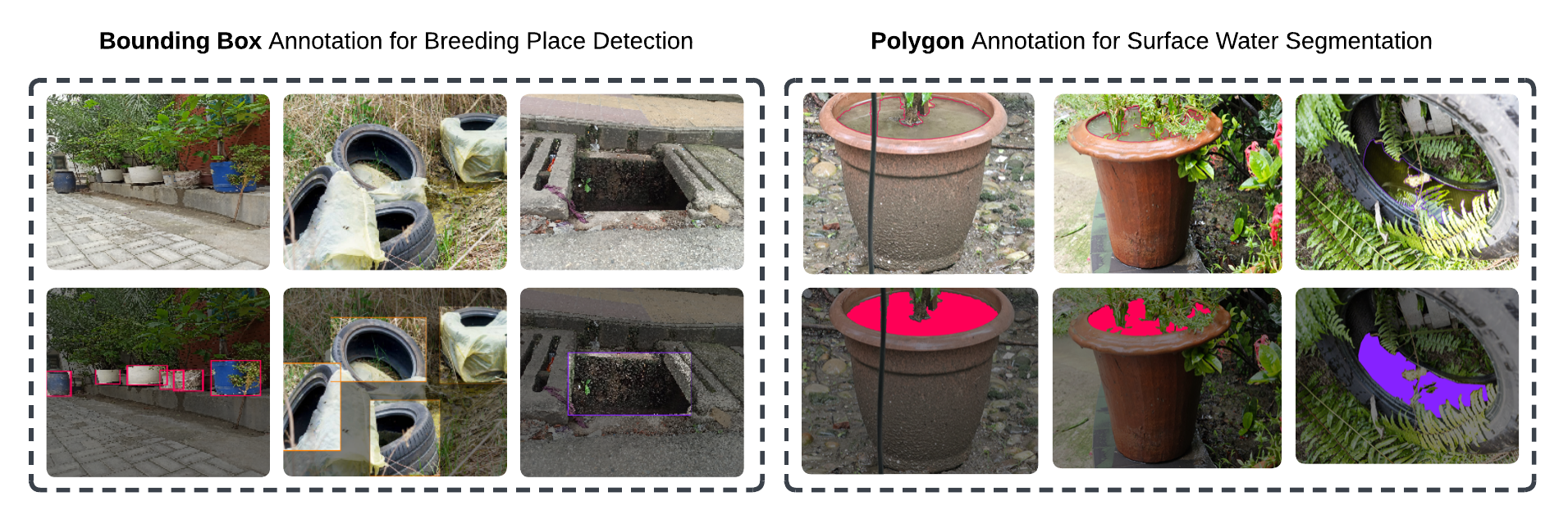}
    \caption{Fully tagged and labeled images for breeding site detection and surface water segmentation}
    \label{fig:annotate}
\end{figure*}

\subsection{Data Collection}

The data collection process is designed to ensure diversity, accuracy, and real-world relevance in capturing mosquito breeding sites and water surfaces. High-quality images are collected from various regions across Bangladesh, covering diverse breeding habitats under both daylight (8 AM–5 PM) and nighttime conditions to enhance dataset variability. To improve model generalization, multiple images are taken from different angles and distances (1–3 meters), ensuring a detailed visual representation. Both natural and artificial breeding sites are documented, though challenges such as unpredictable weather and difficult terrain occasionally impacted data collection. Ethical considerations are prioritized by obtaining permission from local authorities and property owners. The process remains non-invasive, avoiding harm to natural habitats or disruptions to local communities. Anonymization techniques are applied to protect sensitive location details. 

The initial dataset comprises 1,828 images with 3,752 annotations for breeding place detection and 142 images with 253 annotations for water surface segmentation. This ensures a diverse, comprehensive foundation for training models on mosquito habitat detection and surface segmentation. Additionally, a text modality is included in the dataset to enable multimodal analysis. This dataset contains 3,762 instances, each associated with an image and annotated with three text fields: \textbf{(a) Question:} A binary question asking whether the image shows a mosquito breeding site. \textbf{(b) Response:} A 'Yes' or 'No' answer (3,748 \textit{``Yes''} and 14 \textit{``No''} responses). \textbf{(c) Textual Explanation:} A short free-text explanation justifying the response. The average length of the textual explanation statements is approximately 36 tokens. This textual annotation provides semantic context and interpretability, significantly enhancing the dataset's capacity for explainable AI.

\subsection{Data Preprocessing}

The data preprocessing involves annotation, transformations, and augmentation to enhance the dataset. All images are manually annotated using the Roboflow \cite{R1} platform, ensuring precise labeling of mosquito breeding sites and water surfaces. Figure \ref{fig:annotate} shows examples of the annotated images. The following preprocessing steps are applied to each image: \textbf{(a) Auto-Orient:} Images are auto-oriented to correct any device orientation inconsistencies. \textbf{(b) Resize:} All images are resized to 640x640 pixels for uniform input shape. \textbf{(c) Auto-Adjust Contrast:} The contrast is automatically adjusted to enhance visual clarity. To improve model robustness, augmentation techniques are applied: \textbf{(a) Flip:} Horizontal flips doubled the dataset size by varying object orientations. \textbf{(b) Rotation:} Random rotations introduce alignment variations. \textbf{(c) Brightness Adjustment:} Image brightness is varied to simulate real-world lighting conditions. As a result of these augmentations, the total number of images in the dataset increases to 4,425 for the detection part and 331 for the segmentation part. This augmentation strategy significantly enhances the dataset's variability, ensuring that the models trained on this dataset would be more robust and capable of generalizing well to unseen data. For the text modality, the binary responses (\textit{``Yes''} or \textit{``No''}) and the accompanying explanation statements were initially generated using the \texttt{GPT-4o} model. To ensure high-quality annotations, all generated responses are subsequently curated and validated by human annotators. This semi-automated annotation workflow allows efficient dataset expansion while preserving semantic integrity and contextual accuracy. 

\subsection{Distribution Analysis and Folder Structure}

The breeding place detection subset of the dataset comprises a total of 1,828 images with 3,752 annotations distributed across five classes. The selected object categories were derived from existing literature on mosquito ecology and vector-control studies, as well as environmental field observations identifying common stagnant-water containers in endemic regions \cite{dejenie2011characterization, fillinger2009identifying}. The class-wise distribution indicates that the \texttt{coconut\_exocarp} class has the highest number of instances with 923 annotations, followed closely by the \texttt{vase} class with 911 annotations. The \texttt{tire} class contains 780 annotations, while the \texttt{drain\_inlet} and \texttt{bottle} classes have 585 and 553 annotations, respectively. For the segmentation part of the dataset, there are 142 images with a total of 253 annotations across two classes: \texttt{vase\_with\_water} and \texttt{tire\_with\_water}. The \texttt{vase\_with\_water} class has a significantly higher number of annotations, with 181 instances, compared to the \texttt{tire\_with\_water} class, which contains 72 annotations. Table~\ref{tab:annotation_distribution} summarizes the class-wise annotation distribution in our dataset.


\begin{table}[t]
    \centering
    \small
    \renewcommand{\arraystretch}{1.15}
    \begin{tabular}{p{2.6cm}|p{2.6cm}|p{1.6cm}}
        \hline
        \rowcolor{gray!20}
        \textbf{Category} & \textbf{Class} & \textbf{\# Anno.} \\
        \hline
        \multirow{5}{*}{\parbox{2.6cm}{\centering Breeding Place Detection}} 
            & \cellcolor{green!15} coconut\_exocarp   & \cellcolor{green!15} 923 \\
            & \cellcolor{green!5} vase               & \cellcolor{green!5} 911 \\
            & \cellcolor{green!15} tire               & \cellcolor{green!15} 780 \\
            & \cellcolor{green!5} drain-inlet        & \cellcolor{green!5} 585 \\
            & \cellcolor{green!15} bottle             & \cellcolor{green!15} 553 \\
        \hline
        \multirow{2}{*}{\parbox{2.6cm}{\centering Water Surface Segmentation}}
            & \cellcolor{yellow!5} vase\_with\_water   & \cellcolor{yellow!5} 181 \\
            & \cellcolor{yellow!15} tire\_with\_water   & \cellcolor{yellow!15} 72 \\
        \hline
    \end{tabular}
    \caption{Annotation Distribution in the Dataset}
    \label{tab:annotation_distribution}
\end{table}

\begin{figure}[t]
    \centering
    \includegraphics[width=\columnwidth]{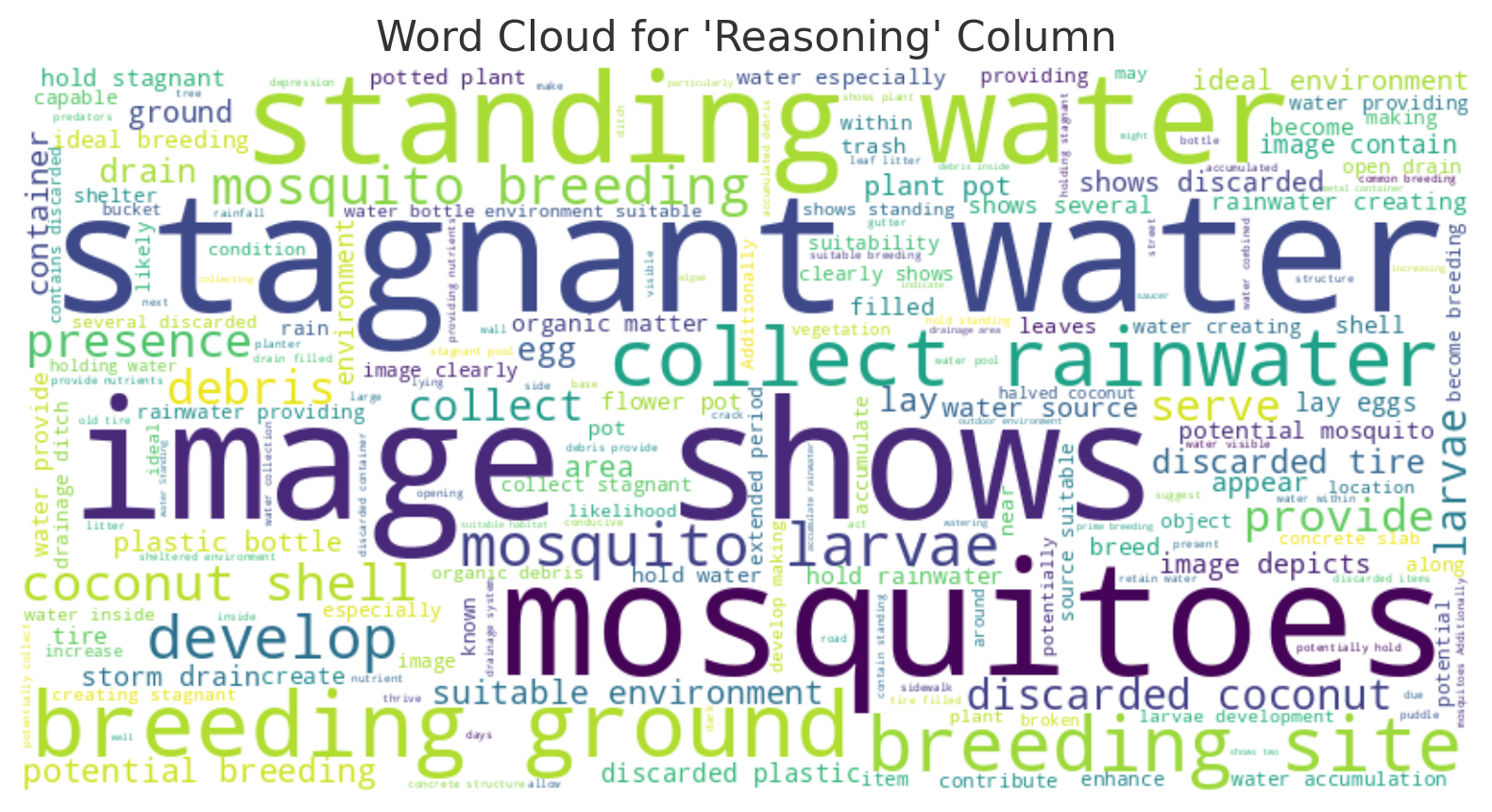}
    \caption{Word Cloud of Common Terms in Explanation Texts}
    \label{fig:wordcloud}
\end{figure}

In addition to visual data, the dataset contains textual annotations in the form of explanation responses that describe the rationale behind each detection or segmentation. Analysis of the explanation texts reveals that the average length is approximately 230 characters, with most entries ranging between 175 and 280 characters. The text lengths follow a roughly normal distribution, indicating consistency in the annotation style. Most frequently occurring terms in the explanation responses include phrases such as “stagnant water,” “mosquitoes,” “mosquito larvae,” and “potential breeding site,” reflecting common descriptors and domain-specific language used during the annotation process. A word cloud is generated to visualize the most frequently occurring terms in the explanation responses, highlighting key descriptors and domain-specific language used during the annotation process. Figure \ref{fig:wordcloud} shows the word cloud derived from the textual explanation data.

The organization of the dataset is designed to optimize data management and accessibility for both object detection and segmentation tasks. The dataset is divided into three main directories: \texttt{Train}, \texttt{Valid}, and \texttt{Test}. Each of these directories contains two sub-folders: \textbf{\texttt{(a) images}}: This folder contains the visual data in the form of images collected from various mosquito breeding sites. \textbf{\texttt{(b) labels}}: This folder contains the corresponding annotation files for each image. The annotations detail the positions and classes of objects or segments identified in the images, serving as a guide for training the machine learning models. In addition to the visual components, the dataset also includes a textual explanation component, which provides natural language justifications for each image annotation. These explanation texts are stored in a separate CSV file that contains a \texttt{filename} column. This column acts as a key to link each textual explanation entry directly to its corresponding image file in the dataset.

\section{Experimental Setup}
\label{sec:methodology}

\begin{figure*}[t!]
    \centering
    \includegraphics[width=\textwidth]{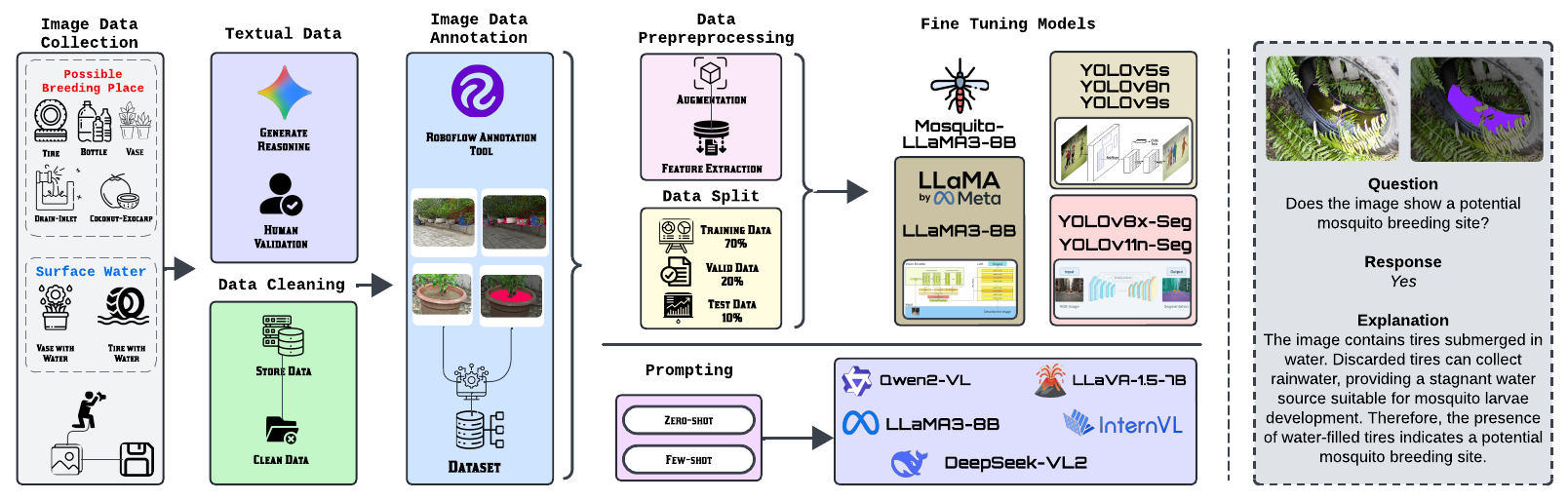}
    \caption{Overview of the \textsc{VisText-Mosquito} pipeline, which involves several key steps: data collection, annotation, preprocessing, and model training. It supports object detection, water surface segmentation, and explanation generation for mosquito breeding site analysis.}
    \label{fig:methodology}
\end{figure*}

In this section, we discuss the detailed experimental setup of our proposed solution shown in Figure \ref{fig:methodology}.

\subsection{Hardware Configurations}

All images were manually reviewed to ensure no personally identifiable information was present. The dataset was split into 70\% training, 20\% validation, and 10\% testing. For object detection and segmentation, models were trained for 100 epochs on a Linux workstation with Ubuntu 24.04, equipped with an NVIDIA RTX 3090 Ti GPU (24 GB) and an AMD Ryzen 5800X CPU. Images were resized to 640×640 pixels for consistency across experiments. 


\subsection{Models}

For the breeding site detection task, we use state-of-the-art object detection models: \texttt{YOLOv5s}, \texttt{YOLOv8n}, \texttt{YOLOv9s}, and \texttt{RT-DETR ResNet-101}. These models are chosen for their lightweight architecture, fast inference, and competitive accuracy, especially for detecting small, object-like containers or wet surfaces associated with mosquito breeding. In parallel, we apply \texttt{YOLOv8x-Seg}, \texttt{YOLOv11n-Seg}, and \texttt{Mask R-CNN (ResNet-101 FPN)} for the water surface segmentation task. These models extend object detection with dense pixel-wise segmentation capabilities, enabling precise boundary-level understanding of water-holding structures such as tires and vases. All models are initialized with pre-trained weights from the COCO dataset and then fine-tuned on the \textsc{VisText-Mosquito} dataset to adapt to our specific domain. To support multimodal understanding, we explore a wide range of large vision-language models for the task of generating natural language explanations based on images. These models are tasked with answering a yes/no question and generating free-form textual justifications grounded in visual evidence. We evaluate both open-source and closed-source VLMs. From the closed-source domain, we test \texttt{Gemini-2.5-Flash}, which is a proprietary model. For open-source alternatives, we evaluate \texttt{LLaVA-1.5-7B}, \texttt{LLaMA3-8B-Vision}, \texttt{DeepSeek-VL2}, \texttt{InternVL-4B}, and \texttt{Qwen2-VL}.

\subsection{Prompting Strategies}

Each vision-language model is tested under two prompting strategies: zero-shot and few-shot. In the zero-shot setting, the model is given only the input image and question, without any in-context examples. This setup evaluates the model's out-of-the-box generalization capacity to unseen prompts. In the few-shot setting, we include three image-question-explanation pairs as exemplars in the prompt to guide the model.

\subsection{Supervised Fine-Tuning}

Beyond prompting, we perform supervised fine-tuning using the \texttt{LLaMA3-8B-Vision} model on the \textsc{VisText-Mosquito} dataset, which is named \texttt{Mosquito-LLaMA3-8B}. During fine-tuning, the model learns to generate explanation text conditioned on the image content and question prompt using a cross-entropy loss over token probabilities. The model is optimized using the AdamW optimizer with a learning rate of \(5 \times 10^{-5}\) and trained using a batch size of 16. Training proceeds for 3 epochs. This process enables the model to capture domain-specific visual-textual associations and produce reliable and interpretable textual explanations.

\subsection{Evaluation Metrics}

To measure the effectiveness of our models, we adopt both vision-based and language-based evaluation metrics. For object detection and segmentation tasks, we report precision, recall, and mean Average Precision at Intersection over Union (IoU) threshold 0.5 (mAP@50). These metrics provide insight into how accurately the models detect and localize breeding sites and water surfaces. For the textual explanation generation task, we use a combination of BLEU, ROUGE-L, and BERTScore to evaluate the quality of generated text. BLEU captures n-gram overlap with reference textual explanation, ROUGE-L assesses longest common subsequences to reflect fluency and relevance, while BERTScore leverages contextual embeddings to capture semantic similarity. 

\section{Result Analysis}
\label{sec:result_analysis}

\begin{figure*}[htbp]
    \centering
    \includegraphics[width=0.95\linewidth]{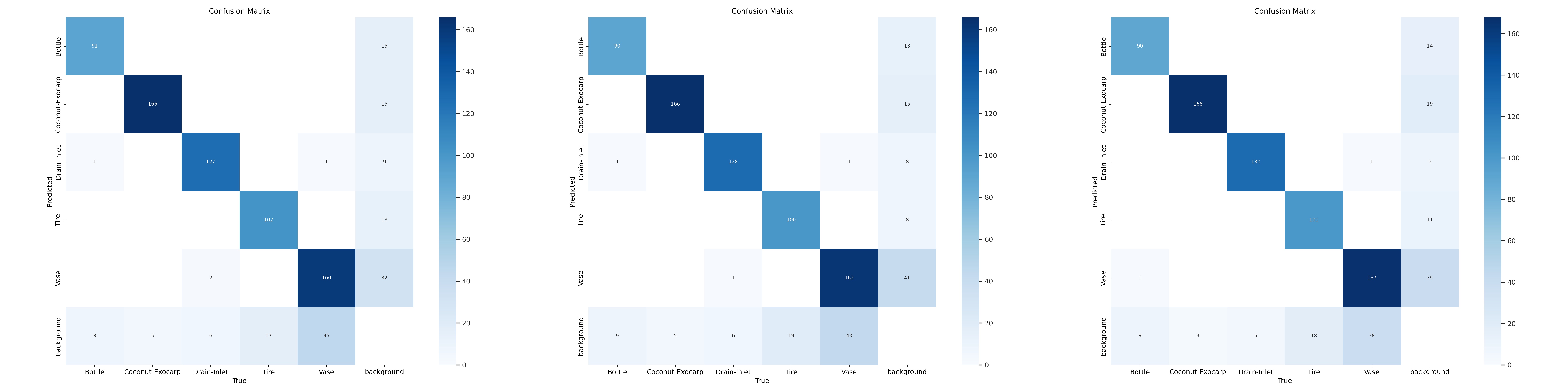}
    
    \vspace{1em}

    \includegraphics[width=0.6\linewidth]{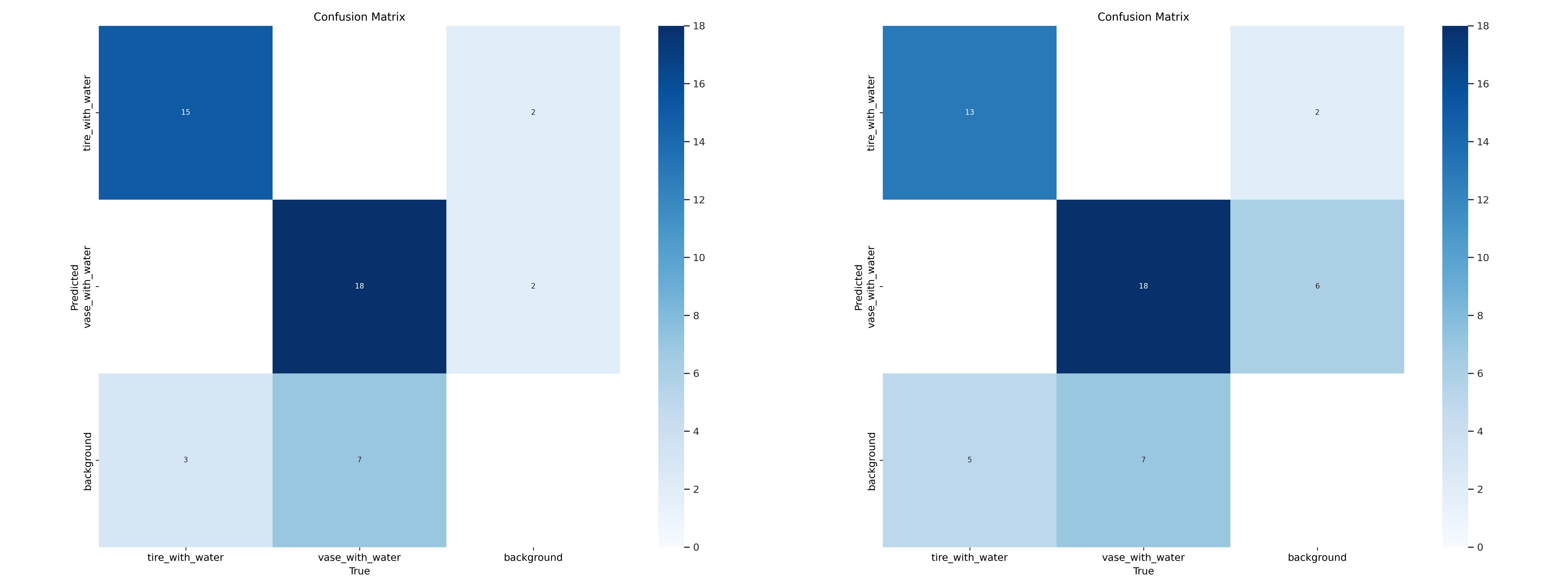}

    \caption{Confusion matrices for (Top) object detection and (Bottom) segmentation tasks.}
    \label{fig:confusion_matrices}
\end{figure*}



\subsection{Object Detection Performance}


To detect potential mosquito breeding sites, we fine-tune and evaluate four object detection models, \texttt{YOLOv5s}, \texttt{YOLOv8n}, \texttt{YOLOv9s}, and \texttt{RT-DETR ResNet-101}, on 1,828 annotated images across five object classes. The detailed performance is summarized in Table~\ref{tab:object_detection_results}.

The \texttt{YOLOv9s} model achieves the highest precision of 0.92926 and mAP@50 of 0.92891, demonstrating its superior ability to accurately localize and classify breeding-related objects. In contrast, \texttt{YOLOv5s} offers the most balanced performance, with a strong precision of 0.91514 and the highest recall of 0.87595. This makes it particularly well-suited for applications where minimizing false negatives is essential, such as early-stage mosquito surveillance. The \texttt{RT-DETR ResNet-101} model achieves a precision of 0.88935, recall of 0.82778, and mAP@50 of 0.87871, showing competitive detection capability but lower overall performance than the YOLO-based variants on this dataset. Meanwhile, \texttt{YOLOv8n} shows slightly stronger recall and mAP@50 than RT-DETR, with a precision of 0.89028, recall of 0.87314, and mAP@50 of 0.90817. These results suggest that the architectural improvements in \texttt{YOLOv9s} contribute to meaningful gains in real-world breeding site detection tasks.



\begin{table}[t]
    \centering
    \small
    \renewcommand{\arraystretch}{1.15}
    \begin{tabular}{c|c|c|c}
        \hline
        \rowcolor{gray!20}
        \textbf{Model} & \textbf{Precision} & \textbf{Recall} & \textbf{mAP@50} \\
        \hline
        \rowcolor{green!15}
        YOLOv5s & 0.91514 & \textbf{0.87595} & 0.92400 \\
        \rowcolor{green!5}
        YOLOv8n & 0.89028 & 0.87314 & 0.90817 \\
        \rowcolor{green!15}
        YOLOv9s & \textbf{0.92926} & 0.86011 & \textbf{0.92891} \\
        \rowcolor{green!5}
        RT-DETR ResNet-101 & 0.88935 & 0.82778 & 0.87871 \\
        \hline
    \end{tabular}
    \caption{Object Detection Model Performance}
    \label{tab:object_detection_results}
\end{table}

\subsection{Water Surface Segmentation Performance}

The segmentation task is focused on identifying water surfaces in objects like vases and tires, critical indicators of mosquito breeding potential. We fine-tuned three advanced models, \texttt{YOLOv8x-Seg}, \texttt{YOLOv11n-Seg}, and \texttt{Mask R-CNN (ResNet-101 FPN)}, on 142 images annotated for \texttt{vase\_with\_water} and \texttt{tire\_with\_water}. Table~\ref{tab:segmentation_results} presents their performance. 

\texttt{YOLOv11n-Seg} consistently achieves the strongest overall performance across all three evaluation metrics. Specifically, it attains the highest precision of 0.91587, compared to 0.89372 for \texttt{YOLOv8x-Seg} and 0.90142 for \texttt{Mask R-CNN (ResNet-101 FPN)}. It also records the best recall of 0.77201, indicating a stronger ability to capture true positive water-surface regions without missing relevant segments. In terms of mAP@50, \texttt{YOLOv11n-Seg} again performs best with 0.79795, while \texttt{Mask R-CNN} achieves a competitive score of 0.79511, slightly surpassing \texttt{YOLOv8x-Seg}.

The inclusion of \texttt{Mask R-CNN} provides a strong two-stage segmentation baseline for comparison with the YOLO-based one-stage segmentation models. Its competitive precision and mAP@50 suggest that region-based segmentation remains effective for capturing structured water-holding regions such as vase interiors and tire cavities. However, \texttt{YOLOv11n-Seg} demonstrates the best balance between localization quality and segmentation consistency, making it the most robust model for pixel-level mosquito habitat identification in visually challenging real-world environments with clutter, shadows, and partial occlusion.



\begin{table}[t]
    \centering
    \small
    \renewcommand{\arraystretch}{1.15}
    \begin{tabular}{c|c|c|c}
        \hline
        \rowcolor{gray!20}
        \textbf{Model} & \textbf{Precision} & \textbf{Recall} & \textbf{mAP@50} \\
        \hline
        \rowcolor{yellow!15}
        YOLOv8x-Seg & 0.89372 & 0.73074 & 0.79345 \\
        \rowcolor{yellow!5}
        YOLOv11n-Seg & \textbf{0.91587} & \textbf{0.77201} & \textbf{0.79795} \\
        \rowcolor{yellow!15}
        Mask R-CNN & 0.90142 & 0.74863 & 0.79511 \\
        \hline
    \end{tabular}
    \caption{Segmentation Model Performance}
    \label{tab:segmentation_results}
\end{table}

\begin{table*}[t!]
\centering
\small
\renewcommand{\arraystretch}{1.15}
\begin{tabular}{l|ccc|ccc}
\hline
\multirow{2}{*}{\textbf{Model}} &
\multicolumn{3}{c|}{\cellcolor{gray!15}\textbf{Zero-shot}} &
\multicolumn{3}{c}{\textbf{\cellcolor{gray!15} Few-shot}} \\
\cline{2-7}
 & \cellcolor{gray!15} BLEU & \cellcolor{gray!15} BERTScore & \cellcolor{gray!15} ROUGE-L & \cellcolor{gray!15} BLEU & \cellcolor{gray!15} BERTScore & \cellcolor{gray!15} ROUGE-L \\
\hline
\rowcolor{gray!15}
\multicolumn{7}{l}{\textbf{Open-source}} \\
\hline
LLaVA-1.5-7B
 & \cellcolor{blue!15}38.2 & \cellcolor{blue!15}0.66 & \cellcolor{blue!15}0.65
 & \cellcolor{blue!15}39.1 & \cellcolor{blue!15}0.68 & \cellcolor{blue!15}0.69 \\
LLaMA3-8B-Vision
 & \cellcolor{blue!5}40.2 & \cellcolor{blue!5}0.71 & \cellcolor{blue!5}0.68
 & \cellcolor{blue!5}42.8 & \cellcolor{blue!5}0.78 & \cellcolor{blue!5}0.72 \\
DeepSeek-VL2
 & \cellcolor{blue!15}36.7 & \cellcolor{blue!15}0.64 & \cellcolor{blue!15}0.63
 & \cellcolor{blue!15}38.2 & \cellcolor{blue!15}0.73 & \cellcolor{blue!15}0.66 \\
InternVL-4B
 & \cellcolor{blue!5}33.5 & \cellcolor{blue!5}0.52 & \cellcolor{blue!5}0.57
 & \cellcolor{blue!5}35.4 & \cellcolor{blue!5}0.56 & \cellcolor{blue!5}0.60 \\
Qwen2-VL
 & \cellcolor{blue!15}31.6 & \cellcolor{blue!15}0.55 & \cellcolor{blue!15}0.60
 & \cellcolor{blue!15}29.3 & \cellcolor{blue!15}0.50 & \cellcolor{blue!15}0.55 \\
\hline
\rowcolor{gray!15}
\multicolumn{7}{l}{\textbf{Closed-source}} \\
\hline
Gemini-2.5-Flash
 & \cellcolor{blue!5}45.9 & \cellcolor{blue!5}0.78 & \cellcolor{blue!5}0.73
 & \cellcolor{blue!5}47.5 & \cellcolor{blue!5}0.83 & \cellcolor{blue!5}0.77 \\
GPT-4o mini
 & \cellcolor{blue!15}43.1 & \cellcolor{blue!15}0.80 & \cellcolor{blue!15}0.75
 & \cellcolor{blue!15}46.2 & \cellcolor{blue!15}0.85 & \cellcolor{blue!15}0.79 \\
\hline
\rowcolor{gray!15}
\multicolumn{7}{l}{\textbf{Fine-tuned}} \\
\hline
\cline{2-7}
& \multicolumn{2}{c}{\cellcolor{gray!15} BLEU} & \multicolumn{2}{c}{\cellcolor{gray!15} BERTScore} & \multicolumn{2}{c}{\cellcolor{gray!15} ROUGE-L} \\
\hline
\textbf{Mosquito-LLaMA3-8B (Ours)}
 & \multicolumn{2}{c}{\cellcolor{blue!15}\textbf{54.7}}
 & \multicolumn{2}{c}{\cellcolor{blue!15}\textbf{0.91}}
 & \multicolumn{2}{c}{\cellcolor{blue!15}\textbf{0.85}} \\
\hline
\end{tabular}
\caption{Zero-shot, Few-shot, and Fine-tuned Textual Explanation Performance Across Vision--Language Models. The fine-tuned model is evaluated only after task-specific adaptation.}
\label{tab:zerofew_results}
\end{table*}


\subsection{Multimodal Explanation Performance}


For the textual explanation task, we fine-tune the \texttt{LLaMA3-8B-Vision} model using image-question-explanation triplets curated from the \textsc{VisText-Mosquito} dataset. The model is trained to generate coherent and semantically grounded justifications that explain whether an image depicts a mosquito breeding site and why. During training, it learns to associate domain-specific visual patterns, such as standing water in tires, vases, or drains, with meaningful language that reflects expert-level explanation. After three epochs, the model converges with a final training loss of 0.0028, suggesting stable training and strong multimodal alignment. To evaluate the generated justifications, we adopt standard natural language generation metrics including BLEU, BERTScore, and ROUGE-L. Our fine-tuned model, referred to as \texttt{Mosquito-LLaMA3-8B}, achieves a BLEU score of 54.7, a BERTScore of 0.91, and a ROUGE-L of 0.85. These results indicate strong n-gram precision, semantic similarity, and structural alignment with human-written reference texts. Qualitative inspection further supports these findings: the model effectively integrates visual context, such as the presence of water, container types, and environmental cues, into fluent, human-readable rationales. Beyond supervised fine-tuning, we assess the generalization and adaptation abilities of other vision-language models under both zero-shot and few-shot prompting setups. In the zero-shot setting, models receive only an image and a prompt without any additional context. In the few-shot setting, each model is shown three image-question-explanation examples before generating responses for unseen inputs. This experimental design enables us to evaluate both generalization from pretraining and responsiveness to in-context learning. Table~\ref{tab:zerofew_results} summarizes the results. Overall, few-shot prompting improved performance across most models, with notable increases in BLEU, BERTScore, and ROUGE-L. For instance, \texttt{LLaMA3-8B-Vision} improved from 40.2 to 42.8 in BLEU, and from 0.71 to 0.78 in BERTScore. \texttt{LLaVA-1.5-7B} and \texttt{DeepSeek-VL2} showed modest but consistent improvements across all metrics, while even the smaller \texttt{InternVL-4B} gained from few-shot examples, despite its relatively lower baseline. Interestingly, the closed-source \texttt{Gemini-2.5-Flash} displayed only minor improvements, possibly due to its already strong zero-shot capabilities limiting headroom for further gain. A notable exception to this trend is \texttt{Qwen2-VL}, which experienced a performance decline in all three metrics under few-shot prompting, dropping from 31.6 to 29.3 in BLEU, from 0.55 to 0.50 in BERTScore, and from 0.60 to 0.55 in ROUGE-L. This suggests that Qwen2-VL may be particularly sensitive to prompt construction or lacks sufficient in-context learning capabilities for domain-specific textual explanation tasks. In contrast, our \texttt{Mosquito-LLaMA3-8B} model exhibits significant improvements across all metrics. Compared to its own zero-shot variant, it achieves a +13.3 point gain in BLEU, +0.04 in BERTScore, and +0.07 in ROUGE-L. These results highlight the clear advantage of supervised domain-specific fine-tuning over reliance on general pretraining or prompt-based learning. Overall, our findings indicate that while few-shot prompting enhances textual explanation capabilities in most models, fine-tuning remains essential for achieving high-accuracy, context-sensitive reasoning in public health surveillance applications.


\subsection{Ablation Study}
\label{sec:ablation_study}

\begin{table*}[t]
    \centering
    \small
    \renewcommand{\arraystretch}{1.15}
    \begin{tabular}{l|c|c|c|c}
        \hline
        \rowcolor{gray!20}
        \textbf{Configuration} & \textbf{BLEU} & \textbf{BERTScore} & \textbf{ROUGE-L} & \textbf{Final Loss} \\
        \hline
        \rowcolor{blue!15}
        \textbf{Baseline (3 epochs, full data)}       & \textbf{54.7} & \textbf{0.91} & \textbf{0.85} & \textbf{0.0028} \\
        \hline
        \rowcolor{blue!8}
        1 epoch (underfitting)                        & 46.2 & 0.82 & 0.76 & 0.0085 \\
        \rowcolor{blue!4}
        5 epochs (overfitting)                        & 49.7 & 0.87 & 0.81 & 0.0016 \\
        \rowcolor{blue!8}
        Learning rate $1\times10^{-5}$                & 47.5 & 0.85 & 0.79 & 0.0043 \\
        \rowcolor{blue!4}
        Learning rate $1\times10^{-4}$                & 48.8 & 0.86 & 0.80 & 0.0039 \\
        \rowcolor{blue!8}
        50\% training examples                        & 42.1 & 0.76 & 0.72 & 0.0071 \\
        \rowcolor{blue!4}
        75\% training examples                        & 48.3 & 0.84 & 0.77 & 0.0052 \\
        \rowcolor{blue!8}
        No image augmentation                         & 46.8 & 0.81 & 0.75 & 0.0060 \\
        \hline
    \end{tabular}
    \caption{Ablation Study on Fine-Tuned \texttt{Mosquito-LLaMA3-8B}}
    \label{tab:ablation_study}
\end{table*}

To understand the effect of different training configurations on textual explanation performance, we conduct an ablation study on the fine-tuned \texttt{Mosquito-LLaMA3-8B} model. We vary critical training hyperparameters, including the number of training epochs, learning rate, and the number of image-question-explanation triplets used during training. The goal is to isolate how each factor contributes to the model’s ability to align visual cues with textual explanations. Table~\ref{tab:ablation_study} summarizes the results. The baseline model (our final version) is trained for 3 epochs with a learning rate of $5\times10^{-5}$ and the full dataset. Reducing the number of epochs to 1 or increasing it to 5 both result in slightly degraded performance, likely due to underfitting and overfitting, respectively. Similarly, lowering the learning rate delays convergence and results in weaker semantic alignment. Using only 50\% of the training examples notably degrades all metrics, demonstrating the importance of dataset size for effective domain-specific textual explanation. The ablation confirms that the final configuration strikes a strong balance between generalization and alignment. Notably, the number of training examples and moderate epoch count are the most influential factors. 

\section{Discussion}
\label{sec:discussion}

Despite the strong performance of both detection and explanation models on the \textsc{VisText-Mosquito} dataset, several challenges and limitations remain, particularly in the context of multimodal understanding. Through qualitative analysis, we observed recurring failure cases among VLMs, especially in zero-shot scenarios. For instance, models like \texttt{Qwen2-VL} and \texttt{InternVL-4B} often failed to generate grounded explanations when visual cues were subtle or occluded. A common example includes misinterpreting a dry tire partially filled with leaves as a mosquito breeding site, where the model incorrectly inferred the presence of water based on contextual noise rather than actual visual evidence. Another frequent failure pattern emerged when visually similar objects led to semantic confusion. VLMs sometimes misclassified objects such as a decorative vase or a plastic bucket as breeding containers even when they lacked water. This suggests that models are often over-reliant on coarse object shape and context rather than finer visual-textual alignment.

Additionally, models struggled when backgrounds were cluttered or when key cues like water surfaces were not prominently visible. These cases highlight the limitations of current VLMs in precisely localizing fine-grained visual patterns critical to public health applications. The disparity between zero-shot and few-shot performance across models further reinforces the importance of context-aware prompting. Closed-source models like \texttt{Gemini-2.5-Flash} performed reasonably well even without training, but open-source alternatives significantly improved under few-shot settings, indicating their reliance on task-specific cues. However, even in few-shot setups, hallucinations were observed, and some models generated plausible yet factually incorrect textual explanations, such as inferring mosquito larvae without any visible evidence. These findings emphasize the need for further advancements in VLM robustness, particularly in environments with ambiguous visual signals. Incorporating spatial grounding, multimodal attention refinement, and post-hoc factuality checks could improve reliability. Moreover, continued dataset expansion with hard negative samples and edge-case explanation examples may help fine-tune these models for real-world vector surveillance scenarios.

\section{Conclusion}
\label{sec:conclusion}

The \textsc{VisText-Mosquito} dataset establishes a comprehensive benchmark for automated detection, segmentation, and textual explanation on mosquito breeding sites. Through rigorous experiments, we show that \texttt{YOLOv9s} yields the best detection accuracy, while \texttt{YOLOv11n-Seg} excels in water surface segmentation. Our fine-tuned \texttt{Mosquito-LLaMA3-8B} model shows strong performance in generating coherent justifications, bridging visual cues with textual explanation. These results underscore the potential of integrating vision-language models into public health monitoring frameworks. By combining high-performance object recognition with textual explanation, our dataset paves the way for intelligent, interpretable, and scalable tools to support vector control strategies. Future work will expand the dataset across ecological regions, enhance textual explanation diversity, and explore prompt-adaptive models for localized intervention planning. 
{
    \small
    \bibliographystyle{ieeenat_fullname}
    \bibliography{main}
}


\end{document}